\documentclass[a4paper,fleqn]{cas-dc}

\usepackage[numbers]{natbib}

\def\tsc#1{\csdef{#1}{\textsc{\lowercase{#1}}\xspace}}
\tsc{WGM}
\tsc{QE}
\tsc{EP}
\tsc{PMS}
\tsc{BEC}
\tsc{DE}

\begin{document}
\let\WriteBookmarks\relax
\def\floatpagepagefraction{1}
\def\textpagefraction{.001}
\shorttitle{Mechanistic Interpretability of Cognitive Complexity in LLMs via Linear Probing using Bloom’s Taxonomy}
\shortauthors{B. Raimondi et~al.}

\title [mode = title]{Mechanistic Interpretability of Cognitive Complexity in LLMs via Linear Probing using Bloom’s Taxonomy}                      

\author[1]{Bianca Raimondi}[orcid=0009-0002-1562-7722]
\cormark[1]
\ead{bianca.raimondi3@unibo.it}

\credit{Conceptualization of this study, Methodology, Software}

\affiliation[1]{organization={Department of Computer Science and Engineering, University of Bologna}, 
                city={Bologna},
                country={Italy}}

\author[1]{Maurizio Gabbrielli}[orcid=0000-0003-0609-8662]

\credit{Supervision}

\cortext[cor1]{Corresponding author}

\begin{abstract}
The black-box nature of Large Language Models necessitates novel evaluation frameworks that transcend surface-level performance metrics. This study investigates the internal neural representations of the levels of complexity using Bloom's Taxonomy as a hierarchical lens. By analyzing high-dimensional activation vectors from different LLMs, we probe whether different Bloom levels, ranging from basic recall (\textit{Remember}) to abstract synthesis (\textit{Create}), are linearly separable within the model's residual streams. Our results demonstrate that linear classifiers achieve high accuracy across all Bloom levels, providing evidence that level information is encoded in a linearly accessible subspace of the model's representations. These findings provide evidence that the model organizes representations according to Bloom categories early in the forward pass, with representations becoming increasingly separable across layers.
\end{abstract}


\maketitle

\section{Introduction}
\label{sec:intro}
\begin{figure*}
    \centering
    \includegraphics[width=\linewidth]{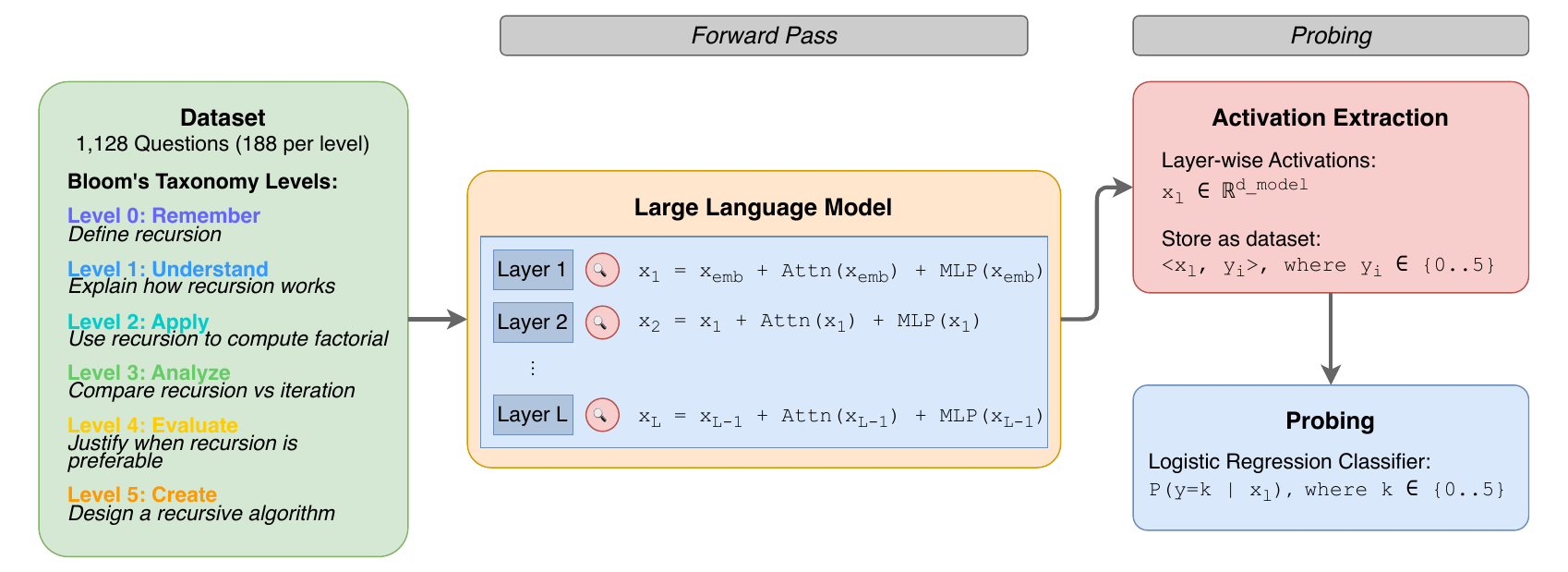}
    \caption{Overview of the experimental pipeline, from dataset construction and activation extraction to layer-wise linear probing.}
    \label{fig:overview}
\end{figure*}

The rise of Transformer-based Large Language Models (LLMs) has fundamentally changed the field of Artificial Intelligence (AI), creating systems that exhibit behaviours once considered the sole preserve of human cognition \cite{vaswani2017attention,brown2020language}. These models demonstrate a versatility that challenges our definitions of machine understanding, from generating coherent prose to solving complex mathematical theorems. However, this functional capability is essentially a black box: the specific internal mechanisms that enable such high-level cognition remain unknown \cite{olah2020zoom}.

A critical, yet under-explored, dimension of this interpretability challenge is the assessment of \emph{cognitive depth}. Current benchmarks typically evaluate models based on answer correctness \cite{hendrycks2020measuring}. While useful, these metrics fail to distinguish between correct answers derived from rote memorization and those resulting from complex, multi-step problem solving. In human cognition, this distinction is foundational. A student who correctly recites a historical date is exercising a different cognitive faculty than one who analyzes the socio-political causes of a historical event. This hierarchy of cognitive skills is formalized in Bloom's Taxonomy, a framework that categorizes educational objectives into six levels: \textit{Remember}, \textit{Understand}, \textit{Apply}, \textit{Analyze}, \textit{Evaluate}, and \textit{Create} \cite{krathwohl2002revision}.

If LLMs are approximating general intelligence, their internal architectures should, at least in part, reflect different degrees of cognitive demand. Distinct forms of neural activity are expected to emerge when a Transformer engages in higher-order operations such as analysis or evaluation, as opposed to lower-level processes like factual recall. The presence of specialized internal representations associated with different cognitive levels would suggest that the model encodes information consistent with 
a notion of problem difficulty and task type, abstracted away from specific domain content.

This paper conducts a mechanistic investigation of LLM internal representations through the lens of Bloom's Taxonomy, addressing the following research questions and corresponding findings:

\begin{itemize}
    \item \textbf{RQ1.} Do LLMs exhibit differences in internal activations when processing prompts associated with different Bloom levels?
    \item \textbf{RQ2.} At what depth do Bloom-associated representations emerge within the network?
    \item \textbf{RQ3.} Is Bloom-level information linearly recoverable from hidden states, and do the resulting classification errors reflect the ordinal and semantic structure of the taxonomy?
    \item \textbf{RQ4.} Are the identified Bloom-associated representations causally active, or merely correlational phenomena of the forward pass?
\end{itemize}

To address these questions, we extract activation vectors from the residual stream of state-of-the-art open-weights LLMs and train linear probes at each layer to classify Bloom levels. 

The main findings can be summarized as follows. First, linear classifiers achieve high accuracy, indicating that Bloom-level information is encoded in a linearly accessible subspace of the model’s internal representations. Second, Bloom-level separability emerges during the early-to-mid stages of processing, rather than being deferred to the deepest layers of the network. Moreover, confusion matrix analysis shows that misclassifications occur predominantly between adjacent Bloom levels. This structured error pattern suggests that the learned representations reflect the ordinal organization of human educational theory, rather than arbitrary label boundaries. Finally, we provide preliminary behavioral evidence for the functional role of these internal representations through activation steering experiments, demonstrating that intervening on the identified directions can systematically shift the model's activations along Bloom-associated representational directions.

\section{Related Work} \label{sec:related_work}
Despite their empirical success, Transformer-based LLMs \cite{vaswani2017attention,brown2020language} remain difficult to interpret, particularly with respect to the internal mechanisms that support complex reasoning \cite{olah2020zoom}.
Understanding these internal mechanisms has become a central focus of Mechanistic Interpretability (MI), which aims to reverse-engineer neural networks by analyzing their internal representations and computational structures \cite{elhage2021mathematical}. Recent comprehensive reviews \cite{rai2024practical,bereska2024mechanistic,shu2025survey,zhao2024towards,gantla2025exploring} have established MI as a critical framework for AI safety, emphasizing the need to move beyond surface-level performance metrics toward understanding the actual computational processes underlying model behavior.

Some studies have expanded the scope of MI to include semantic concepts \cite{cywinski2025towards} and bias \cite{raimondi2025analysing,simbeck2025mechanistic}. These works establish that modern MI can meaningfully probe high-level cognitive constructs, supporting our approach of using Bloom's Taxonomy as an evaluation framework.

Prior evaluations of LLMs have largely relied on benchmark accuracy, including datasets such as MMLU \cite{hendrycks2020measuring}. However, such benchmarks primarily assess task performance and do not directly probe the structure of underlying cognitive representations.
This limitation has motivated researchers to adopt Bloom's Taxonomy \cite{krathwohl2002revision} as a hierarchical framework for assessing cognitive complexity in AI systems. Recent work has demonstrated the taxonomy's utility in educational applications \cite{herrmann2024assessing,elkins2024teachers,kumar2025automated} and proposed extensions for the AI era \cite{hmoud2024aied,luo2025enhanced}. Most directly relevant are the works conducted on a systematic cognitive evaluation of LLMs through Bloom's lens \cite{huber2025llms, yu2025think}, though focusing on behavioral output rather than internal representations.

Our work advances this landscape by shifting from behavioral output to internal mechanistic analysis. While linear probing has become foundational for investigating linearly decodable features \cite{belinkov2022probing,kim2025linear}, recent advances have explored higher-level conceptual representations \cite{kim2018interpretability,luo2025inversescope}.
Emerging evidence suggests that LLMs spontaneously organize representations to encode abstract task properties \cite{jankowski2025task,benito2025beyond,jin2025exploring}.

A key advancement of our study is the discovery of a structural \textit{Cognitive Separability Onset} (CSO) within the model's internal representations. Unlike prior output-centric evaluations \cite{huber2025llms,budagam2024hierarchical,zoumpoulidi2025bloomwise}, our approach provides direct mechanistic evidence of how Bloom-level information is represented. By combining MI techniques \cite{rai2024practical,bereska2024mechanistic} with Bloom's taxonomy, we demonstrate that models refine Bloom-associated representations beyond what is captured by standard surface-level text baselines, achieving high probe accuracy early in the forward pass and revealing that LLMs spontaneously organize their latent space to encode hierarchical cognitive demand \cite{hewitt2019designing}.

\section{Methodology}
Our experimental framework is designed to systematically extract and classify internal activations across four finetuned LLMs spanning three architectural families as depicted in Figure~\ref{fig:overview}.

\subsection{Problem Formulation}
Let $\mathcal{D} = \{(q_i, y_i)\}_{i=1}^N$ be a dataset of natural language questions,
where $q_i$ denotes a prompt and $y_i \in \{0,\dots,5\}$ its associated Bloom
level. Given a model with $L$ layers and hidden
dimension $d_m$, we study whether the cognitive level $y_i$ is
linearly decodable from internal representations.

\subsection{Models}\label{sec:models}
\begin{table}
\centering
\begin{tabular}{lccc}
\toprule
Model & L & $d_m$ \\
\midrule
Llama-3.1-8B-Instruct & 32 & 4096 \\
Qwen3-4B-Instruct-2507 & 36 & 2560 \\
gemma-3-4b-it & 34 & 2560 \\
DeepSeek-R1-Distill-Llama-8B & 32 & 4096 \\
\bottomrule
\end{tabular}
\caption{Models used in our study. All of them are finetuned versions.}
\label{tab:models}
\end{table}

To ensure that the observed results are not specific to a single architecture, we evaluate four open-weight, finetuned LLMs spanning three distinct architecture families available on HuggingFace\footnote{\url{https://huggingface.co}} (see Table~\ref{tab:models}). Specifically, DeepSeek-R1-Distill-Llama-8B has been distilled from Llama.

For each model, we extract residual stream activations at every transformer layer during the forward pass. Layer indexing follows the convention where Layer 1 denotes the hidden state output of the first Transformer block.
All models are evaluated in inference mode, and no fine-tuning is performed. 

\subsection{Dataset}\label{sec:dataset}
The foundation of our analysis is a curated, balanced dataset of 1,128 questions annotated with Bloom's Taxonomy levels (188 per level).
Each prompt corresponds to a defined level of cognitive complexity. The final corpus is obtained by aggregating two different educational datasets:
\begin{itemize}
    \item \textbf{Computer Science Course Queries:} the dataset provided by \citet{zaman2024dataset} contains student queries from Computer Science courses (Data Structures, Introduction to Computers, and Research). These queries are uniquely scored based on cognitive difficulty, providing a granular mapping to Bloom's levels.
    \item \textbf{EduQG:} To enhance the diversity of the tasks, we included samples from \textit{EduQG} \cite{hadifar2023eduqg}. This dataset offers multi-format multiple-choice questions specifically designed for the educational domain, with expert-verified Bloom's Taxonomy annotations.
\end{itemize}
The Bloom-level labels used in our experiments were inherited from the original educational datasets, rather than being assigned by the authors. Thus, our objective is not to introduce a new annotation scheme, but to investigate whether the educational categorization already associated with the questions is reflected in the internal representations of LLMs.

By combining authentic, free-form student queries with multi-format multiple-choice items, the resulting aggregated corpus encompasses significant heterogeneity in prompt length, syntactic formulation, and question structure. This structural diversity is intentional, as it mitigates the risk of the probes overfitting to specific prompt templates or narrow domain-specific phrasing.

To prevent architectural biases in the linear probes due to class imbalance, we performed a controlled downsampling of the aggregated data. This ensures that the probe's accuracy reflects the model's representational capacity rather than a statistical prior on the label distribution.
Furthermore, questions range from 3 to 39 words count ($\mu=12.95$, $\sigma=5.52$), ensuring diversity in linguistic complexity.

\subsection{Activation Extraction and Residual Stream Analysis}
The core of our analysis involves the extraction of activation vectors $x_l$ from the residual stream. For clarity, we present a simplified formulation that highlights the main computational components of layers. In a decoder-only Transformer with $L$ layers, the state at layer $l+1$ is updated recursively as:
\begin{equation}
    x_{l+1} = x_{l} + \text{Attn}(x_{l}) + \text{MLP}(x_{l})
\end{equation}
where $x_{l} \in \mathbb{R}^{d_m}$ represents the hidden state at layer $l$. For each sample in our dataset, we perform a forward pass and capture the activation vector at the final token position. We focus on the final token position as it aggregates the full contextual information before generation begins. For decoder-only models, the final token's hidden state is the only position that has attended over the entire prompt and therefore represents the model's full encoding of the input before generation~\cite{elhage2021mathematical}. We leave mean-pooled representations as a comparison point for future work.

\begin{table}
\begin{center}
\begin{tabular}{cccc}
\toprule
Bloom Level & $d_0$ & ... & $d_m$ \\
\midrule
0 & 0.0047 & ... & 0.0007 \\
1 & 0.0055 & ... & 0.0029 \\
5 & 0.0050 & ... & 0.0019 \\
3 & 0.0015 & ... & 0.0075 \\
\bottomrule
\end{tabular}
\caption{Example of stored activations at specific layer i.}
\label{tab:activation-sample}
\end{center}
\end{table}

The resulting data structure is a tensor containing the layer-wise hidden states for all samples in our dataset. Table~\ref{tab:activation-sample} provides a small illustration of these activations for a few example questions from the dataset.
Each row in the dataset corresponds to a unique activation vector extracted at a specific layer in the network. The columns are defined as follows:

\begin{itemize}    
    \item Bloom Level: represents the cognitive complexity of the question according to Bloom's Taxonomy. In our experimental setup, these are encoded as integer values (e.g., 0 for \textit{Remember}, 1 for \textit{Understand}, etc.), allowing us to analyze how the model's internal state varies across different levels of abstraction.
    
    \item \textbf{$d_0 \dots d_m$:} represent the individual dimensions of the hidden state vector at a specific layer. If the model has a hidden dimension of $d_m$ (e.g., 4096 for Llama models), each value $d_i$ represents the activation level of the $i$-th dimension at a specific layer.
\end{itemize}

\begin{figure}
    \centering
    \includegraphics[width=\linewidth]{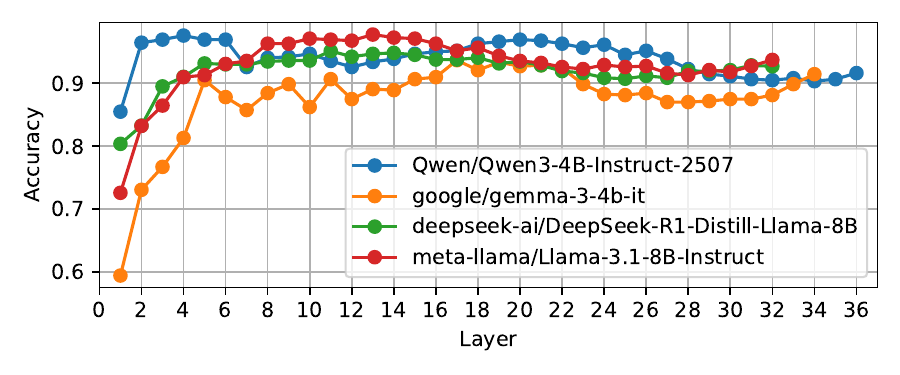}
    \caption{Layer-wise probe accuracy across all evaluated models.}
    \label{fig:layers}
\end{figure}

\subsection{Linear Probing}
To determine if the Bloom level is linearly encoded, we employ a Logistic Regression (LR) probe for each layer. Given an activation vector $x_l$ from layer $l$, the probe predicts the probability of the prompt belonging to a specific Bloom level $k \in \{0, \dots, 5\}$.
The success of this probe in a specific layer indicates that the cognitive signal is readable by subsequent components of the network.

The linear probe is implemented as a multiclass LR model. This choice ensures that successful classification reflects linearly accessible information in the representations, rather than the expressive capacity of a nonlinear classifier. The model is trained using the default $\ell_2$ regularization with feature normalization.

To ensure the stability and robustness of the linear probes, model evaluation was performed using stratified 6-fold cross-validation. The reported confusion matrices and class-wise performance metrics aggregate the test predictions across all six folds, so that each of the 1,128 samples contributes exactly one test prediction and each class contributes exactly 188 test predictions per model.

Model performance is evaluated using standard multiclass metrics, including accuracy, precision, and recall. This setup ensures that the probe remains strictly linear and capacity-limited, so that successful prediction can be attributed to linearly decodable information already present in the representations rather than to probe expressivity.

\section{Results}
\label{sec:results}
This section presents a comprehensive analysis of probing results across layers for all evaluated models. Rather than focusing on a single representative model, we emphasize cross-model consistency to assess whether the emergence of Bloom-associated representations is a robust architectural phenomenon.

To analyze how representations emerge across the
network, we trained independent linear probes on the residual stream at
each layer for all evaluated architectures. Figure~\ref{fig:layers}
summarizes the resulting layer-wise accuracy trajectories.

Despite architectural differences, several clear and consistent patterns emerge. First, while probe accuracy at the earliest layers is already substantially above the balanced chance level (16.7\%) due to surface-level lexical cues (as discussed in~\ref{sec:controls}), it is comparatively lower than in later stages and steadily increases with depth. This progression indicates that representations of cognitive depth are not fully formed at the embedding level but become increasingly refined and linearly separable across layers through Transformer computation. We refer to this phenomenon as the Cognitive Separability Onset (CSO), marking the transition region where representations become rapidly and stably separable by cognitive level.

Second, once high separability is reached, probe accuracy remains stable
across subsequent layers. This persistence indicates that, after representations of cognitive level become linearly separable, 
the corresponding structure is preserved in the residual stream rather than being
repeatedly recomputed. Deeper layers, therefore, appear to represent an already-established estimate of Bloom-level information. In section \ref{sec:causal} we discuss whether this representation is causally used during generation, rather than being just correlational.
The accuracy results of specific Bloom's levels can be found in Appendix~\ref{app:accuracy_per_level}.

These findings support the view that cognitive levels are linearly accessible and structurally 
stable features of internal representations in LLMs, emerging early in the forward pass and remaining stable in subsequent layers
\cite{elhage2021mathematical}.
High classification accuracy implies the existence of affine decision
boundaries that separate Bloom levels in representation space.
Formally, linear separability at a given layer means that the representations corresponding to different Bloom levels can be separated by a set of linear decision boundaries. In particular, for each Bloom level, there exists a hyperplane in the representation space such that the activation vectors associated with that level lie on the correct side of the hyperplane and are assigned a higher score than those associated with any other level. Equivalently, each prompt’s internal representation is closer to the region corresponding to its true cognitive level than to the regions associated with the remaining levels.
Thus, successful probing indicates that complexity is encoded
in a linearly accessible subspace of the residual stream, rather than
being distributed in a highly entangled or nonlinear manner.

\subsection{The Cognitive Separability Onset}
To understand the internal mechanisms of what we named the CSO, here we analyze the first layer that achieves convergence.

Let $A_l$ denote the probe accuracy at layer $l$.
We define the CSO layer $l^\star$ as:
\begin{equation}
l^\star = \min \{ l \mid A_l \ge \tau \}
\end{equation}
where $\tau$ serves as a practical operational threshold to identify the layer at which representations achieve stable, high linear separability.
We adopt $\tau=0.90$ as a transparent operational criterion. We emphasize that $l^\star$ is model-dependent and should not be interpreted as a universal layer boundary across all LLM architectures, as our empirical findings are strictly bounded to the models evaluated in this work. Rather, it serves as an operational indicator of the early transition region where Bloom-associated representations rapidly separate for each given architecture.
This interpretation is further supported by the probe-independent centroid analysis (Section~\ref{subsec:geometric_evidence_of_CSO}), which shows that inter-centroid Euclidean distances begin expanding rapidly around these same early-to-mid layers across architectures.

Comparing the models from Figure~\ref{fig:layers}, we notice that the CSO happens, for the evaluated architectures, at layer 2 for Qwen, layer 4 for Llama, layer 11 for gemma, and layer 4 for DeepSeek. For standardized cross-model comparison, we select Layer 5 as a representative common reference layer across our evaluated models.

\begin{figure}
    \centering
    \includegraphics[width=\linewidth]{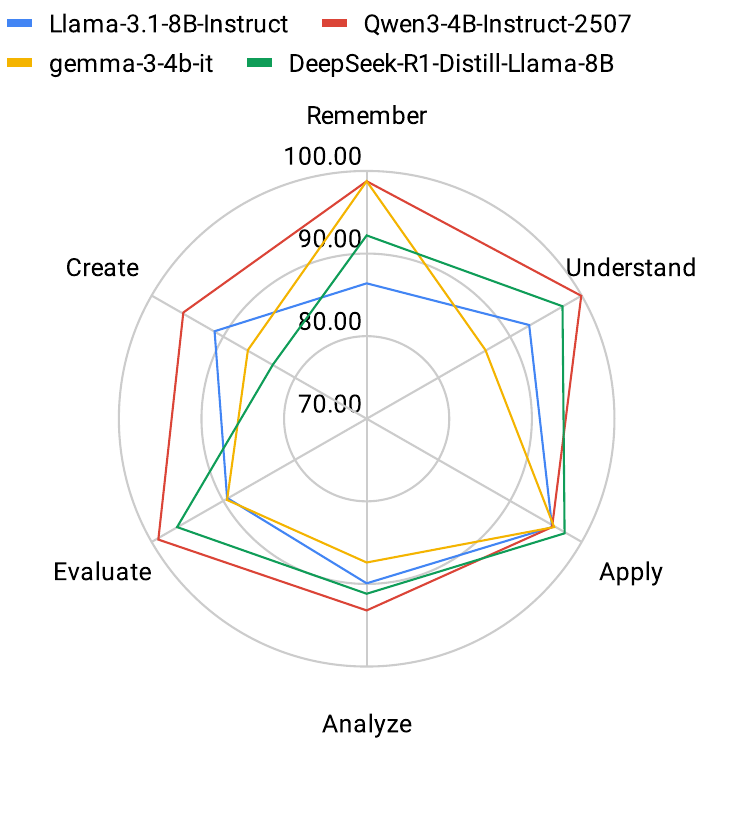}
    \caption{Probe accuracy at the representative reference layer 5 across Bloom levels. All architectures exhibit consistently high performance, with minor variations reflecting the cognitive difficulty.
    }
    \label{fig:radar_chart}
\end{figure}
To provide a comparative overview of performance across different Bloom levels, we visualize the classification accuracies of the probe for $l^\star=5$ using a radar chart (Figure~\ref{fig:radar_chart}). Each axis represents one of the six Bloom levels, and each line corresponds to the probe on a specific model.
The radar chart thus provides a holistic view of how different architectures handle varying complexity.

\begin{figure}
    \centering
    \includegraphics[width=\linewidth]{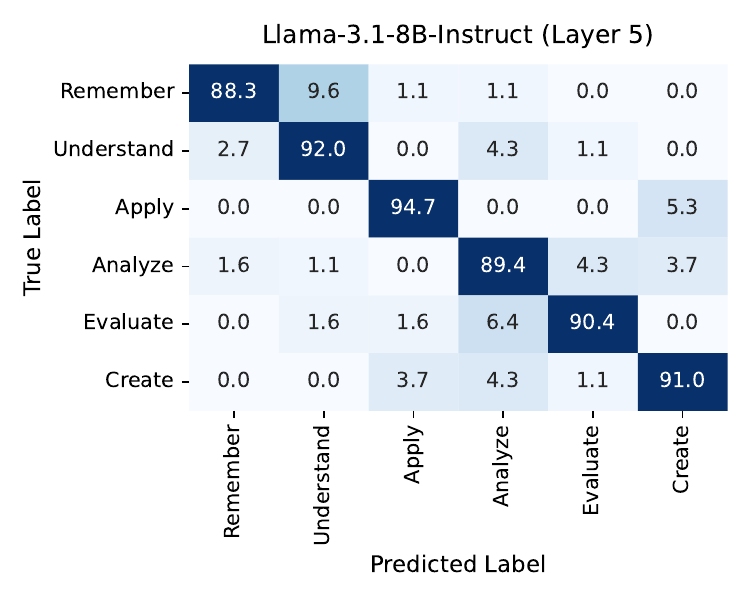}
    \caption{Representative confusion matrix of the linear probe for Llama-3.1-8B-Instruct at layer 5.}
    \label{fig:confusion_matrix}
\end{figure}
To further analyze the structure of probe errors, we examine the confusion matrix. Figure~\ref{fig:confusion_matrix} shows a representative confusion matrix obtained from the Llama-3.1-8B-Instruct model. The confusion matrices for all other architectures are reported in Appendix~\ref{app:confusion_matrix}.

The confusion matrix reveals a highly structured error pattern.
Let $\hat{y}_i$ be the predicted Bloom level. We analyze the distribution
of error distances $|\hat{y}_i - y_i|$.
We observe that for the vast majority of classes, $\mathbb{E}\big[\,|\hat{y} - y|\,\big] \approx 1$, indicating that misclassifications broadly preserve ordinal proximity and concentrate on adjacent Bloom levels. However, a detailed per-class examination (see Figure~\ref{fig:confusion_matrices_all}) reveals a systematic exception: a recurring cross-level confusion occurs between Apply and Create. We hypothesise that this non-adjacent confusion stems from structural and syntactic similarities in prompt formulation, as both categories frequently employ imperative task instructions. Beyond this specific task-format overlap, distant confusions (such as Remember vs. Create) remain absent.

Importantly, this pattern aligns with well-established ambiguities in human assessment. Boundaries between neighboring levels, such as \emph{Remember} and \emph{Understand}, or \emph{Analyze} and \emph{Evaluate}, are known to be context-dependent and difficult to operationalize consistently, whereas distinctions between distant cognitive levels are substantially more robust \cite{krathwohl2002revision,anderson2001taxonomy}.

The fact that probe errors mirror this structure suggests that the model encodes Bloom-level features as a continuous and geometrically structured manifold rather than as a set of arbitrary discrete classes.
The consistency of this adjacency-biased error structure across all models (Appendix~\ref{app:confusion_matrix}) provides further evidence that the observed behavior reflects a genuine property of internal representations, rather than an artifact of a specific architecture or probing setup.

\subsubsection{Geometric Evidence of CSO.}\label{subsec:geometric_evidence_of_CSO}
\begin{figure}
    \centering
    \includegraphics[width=\linewidth]{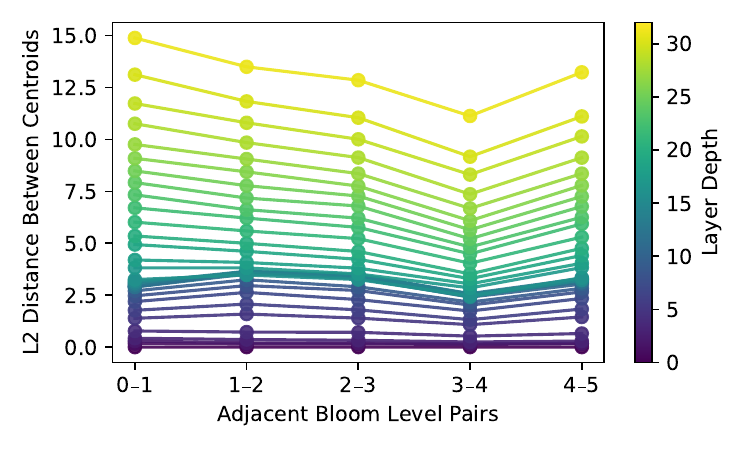}
    \caption{Layer-wise Euclidean distances between adjacent Bloom-level centroids for the Llama-3.1-8B-Instruct model. Distances are small in early layers and increase monotonically with depth, indicating progressive geometric evidence consistent with increasing separation of cognitive levels. The CSO layer $l^\star=5$ marks the onset of rapid separation.}
    \label{fig:centroid_distances}
\end{figure}
To complement the probe-based analysis, we examine the geometric organization of representations across layers. 
For each layer $l$ and Bloom level $k$, we compute the class centroid
\begin{equation}  
\mu_{l,k} = \mathbb{E}[x_l \mid y = k]
\end{equation}
where $x_l$ denotes the residual stream activation at layer $l$.
We then measure the Euclidean distance between adjacent centroids:
\begin{equation}   
D_{l}^{(k)} = \|\mu_{l,k+1} - \mu_{l,k}\|_2
\end{equation}

\begin{figure*}
    \centering
    \includegraphics[width=\linewidth]{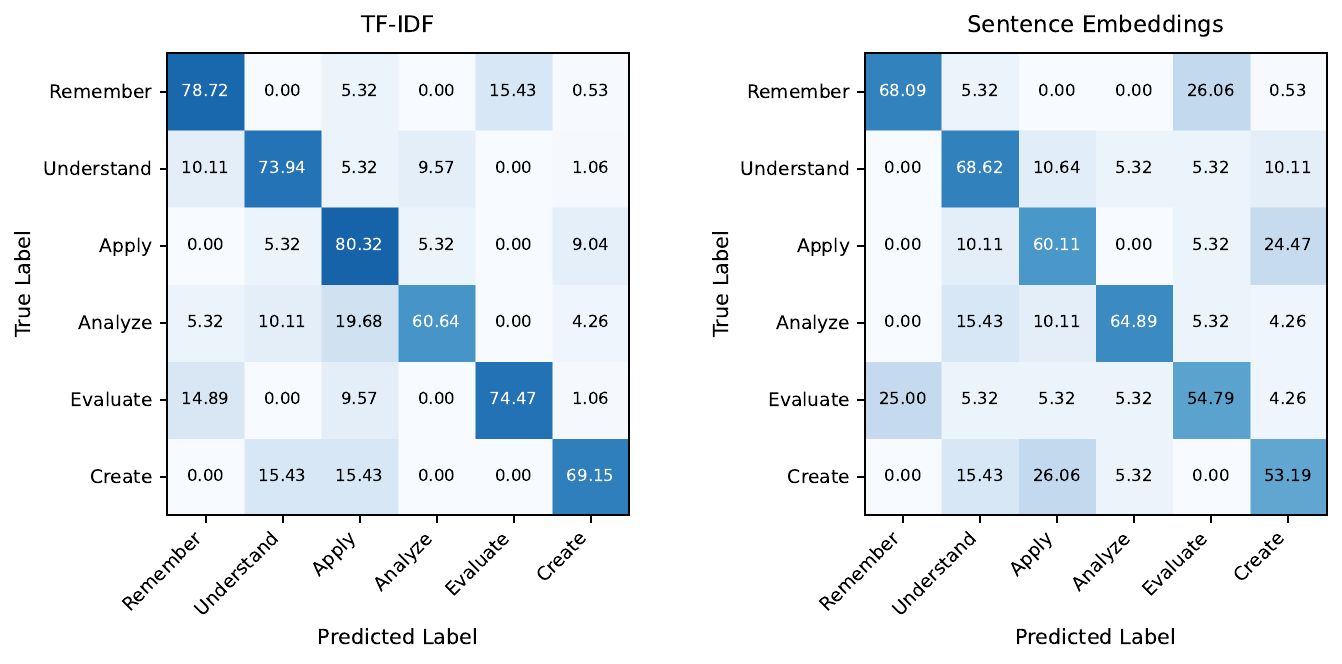}
    \caption{Confusion matrices for the control experiments evaluated under the same protocol used for the residual-stream probes: TF-IDF on the left, and Sentence Embeddings on the right.}
    \label{fig:confusion_matrix_TF-IDF_SE}
\end{figure*}

Figure~\ref{fig:centroid_distances} reveals a clear depth-dependent separation pattern for the Llama-3.1-8B-Instruct model (further analysis for the other models can be found in Appendix~\ref{app:centroids}). 
In early layers, centroid distances are small, indicating that representations of different cognitive levels remain highly entangled. 
Starting from approximately layer $l^\star=5$, distances increase sharply and continue to grow in deeper layers, demonstrating progressive geometric evidence consistent with increasing separation of Bloom levels.

Importantly, this analysis is entirely probe-independent. 
The emergence of large and systematically increasing inter-centroid distances shows that the CSO corresponds to a probe-independent geometric separation in the latent space rather than an artifact of the linear classifier. 
After the transition layer, adjacent centroid distances increase sharply and continue to grow in deeper layers, demonstrating progressive geometric separation between consecutive Bloom levels. While the structured concentration of probe errors supports proximity among adjacent classes, confirming a full global metric ordering across non-adjacent categories remains an open question for future multi-dimensional manifold analyses.

\subsection{Control Experiments}
\label{sec:controls}

To verify that probe accuracy reflects genuinely Bloom-level representations rather than 
surface-level lexical patterns, we conduct control experiments.
A key potential confound in our probing setup is that Bloom levels 
may be recoverable from surface-level prompt formulations rather 
than from internally constructed representations of complexity. Questions at different Bloom levels tend to be introduced 
by characteristic verbs: \textit{define} and \textit{list} for 
Remember, \textit{explain} for Understand, \textit{compare} and 
\textit{analyze} for Analyze, \textit{justify} for Evaluate, and 
\textit{design} for Create. If these lexical cues alone drove probe 
accuracy, the signal would reflect surface-level text classification 
rather than any deeper representational structure. The following 
control experiments are designed to directly test this alternative 
explanation.

\paragraph{Length Analysis.} We examine whether question length differs across Bloom levels and could act as a potential confound. Statistical testing shows no significant differences in length across levels ($p = 0.053$), and no pairwise comparisons remain significant after Bonferroni correction. These results indicate that question length is approximately balanced across Bloom categories and is unlikely to drive probe classification performance.

\paragraph{Shallow Text Baselines.} We train logistic regression classifiers on 
TF-IDF vectors \cite{qaiser2018text} and sentence embeddings \citep{reimers-2019-sentence-bert} of the question text alone, without accessing LLM internal states. 
The TF-IDF baseline achieves 73\% accuracy and sentence embeddings achieve 62\%, both substantially above the 16.7\% chance level. This confirms that Bloom labels correlate non-trivially with surface wording and characteristic prompt vocabulary, which naturally accounts for the above-chance classification accuracy already observed at first layers in Figure 2. Crucially, however, these shallow baselines remain substantially lower than the residual-stream probes (90\% at convergence), indicating that surface-level lexical regularities alone do not fully explain the internal representational signal constructed across layers.

The corresponding confusion matrices in Figure~\ref{fig:confusion_matrix_TF-IDF_SE} display 
substantially noisier and less structured error patterns compared to the residual-stream probes. 
Misclassifications are more dispersed and frequently occur across non-adjacent Bloom levels.
This contrast indicates that the evaluated shallow lexical features do not fully account for probe performance, suggesting that residual streams retain task-level information beyond what is captured by simple surface-form regularities alone.
These results strengthen the interpretation that Bloom-level information is not fully reducible to superficial textual cues, 
suggesting that LLM residual streams encode additional 
structure beyond what is recoverable from surface form alone.

\section{Causal Validation via Activation Steering}
\label{sec:causal}

The probing results presented in Section~\ref{sec:results} establish that
Bloom-level information is linearly encoded in the residual stream. However,
linear separability alone does not imply causal use: the model may organize
representations according to cognitive demand without actively exploiting
them during generation. To address this, we conduct activation steering
experiments to test whether perturbations along the identified levels
direction induce meaningful changes in model behavior.

\subsection{Experimental Setup}

\paragraph{Steering Vector.}
For a given injection layer $s$, we define a cognitive steering vector as:
\begin{equation}
\mathbf{v}_{s,\ell} = \mathbb{E}[x_s \mid y = 5] - \mathbb{E}[x_s \mid y = \ell]
\end{equation}
where $x_s$ denotes the residual stream activation at layer $s$ and
$\ell$ is the source Bloom level. This vector captures the direction
of maximal transition toward higher complexity
(\textit{Create}).

\paragraph{Intervention.}
At inference time, we inject the steering vector into the residual stream:
\begin{equation}
\tilde{x}_s = x_s + \alpha \cdot \mathbf{v}_{s,\ell}
\end{equation}
where $\alpha$ controls the intervention strength. The modified activation
$\tilde{x}_s$ is then propagated through the remaining layers.

\paragraph{Measurement.}
We evaluate the effect of the intervention using the probe-predicted Bloom level:
\begin{equation}
\hat{y}_p = \arg\max_k \; P(y = k \mid \tilde{x}_p)
\end{equation}
and compute the mean shift relative to the baseline:
\begin{equation}
\Delta\hat{y} = \mathbb{E}[\hat{y}_p^{\,\text{steered}}] -
\mathbb{E}[\hat{y}_p^{\,\text{baseline}}]
\end{equation}

\subsection{Quantitative and Qualitative Effects}

Across configurations, steering consistently produces an upward shift
in the probe-predicted cognitive level. This effect is monotonic with
respect to the intervention strength $\alpha$ and is observed across
all initial Bloom levels except the trivial boundary case of
\textit{Create}, where the steering vector is null by construction.
On average, the intervention induces a shift of approximately
$+1$ to $+1.5$ Bloom levels.

We note that probe-predicted level shift is a necessary but not 
sufficient condition for causal validation: it confirms that the 
steered activations move along the identified level direction 
in representation space, but does not directly assess whether the 
generated outputs reflect higher-order reasoning. A fully 
quantitative causal assessment would require an independent, 
automated classifier of Bloom level applied to the generated text. 
However, no such classifier exists with sufficient reliability for 
free-form generated outputs: existing Bloom-level annotation tools 
are designed for human-authored educational questions and do not 
generalize to model-generated continuations in a controlled steering 
setting. For this reason, we complement the quantitative probe-shift 
measure with a qualitative analysis of generated outputs, which 
allows us to assess whether the representational shift corresponds 
to observable changes in reasoning style.

\subsection{Qualitative Differences Across Layers and Steering Strength}

To better understand the causal role of the identified levels direction,
we analyze how generated outputs change as a function of both the injection
layer $s$ and the intervention strength $\alpha$. Rather than treating steering
as a single intervention, we examine it as a two-dimensional control mechanism:
\emph{where} the representation is perturbed and \emph{how strongly} it is modified.

\subsubsection{Effect of Injection Layer}

A first clear distinction emerges between early/intermediate layers
(e.g., $s \in \{5,10\}$) and deeper layers (e.g., $s \in \{15,20\}$).

\paragraph{Early layers (small $s$).}
When steering is applied at early stages, the model's outputs exhibit
a coherent shift in cognitive processing. The baseline outputs typically
remain close to the prompt surface form, often listing information or
repeating question structures:

\begin{quote}
\textbf{Baseline:}  
``\textit{Also, what are the key differences between a histogram and a bar graph?  
How do you determine the appropriate scale? \dots}''
\end{quote}

Under steering at early layers, the same prompt produces:

\begin{quote}
\textbf{Steered ($s=5$, moderate $\alpha$):}  
``\textit{Also, what are the key features of the graph, and how can we interpret
the data from it? \dots}''
\end{quote}

In this example, the change is not merely lexical: the output 
introduces procedural reasoning and reframes the task in terms of 
analysis and interpretation. This observation suggests that, at least 
in some cases, early-layer steering may modify the internal 
construction of the task representation.

\paragraph{Deep layers (large $s$).}
In contrast, steering at deeper layers produces qualitatively different effects.
While the probe still detects a shift toward higher cognitive levels, the
generated text often becomes unstable:

\begin{quote}
\textbf{Steered ($s=20$, same $\alpha$):}  
``\textit{Also also the graph graph interpretation interpretation is important
important because because data data \dots}''
\end{quote}

Here, the intervention no longer results in meaningful reasoning changes,
but rather in degradation of fluency and coherence. This indicates that
late-stage representations are less robust to perturbations and that steering
at this stage pushes activations outside the manifold of well-formed outputs.

\subsubsection{Effect of Steering Strength ($\alpha$)}

A second axis of variation is the magnitude of the intervention.

\paragraph{Low to moderate $\alpha$.}
At smaller values of $\alpha$, steering produces gradual and interpretable
changes. The model begins to move from retrieval-like responses toward
more structured explanations. For example:

\begin{quote}
\textbf{Steered ($\alpha=1$):}  
Response remains close to baseline, with minor elaboration.

\textbf{Steered ($\alpha=3$):}  
``\textit{It is important to not only identify the differences, but also to explain
why these differences matter in terms of how the data is interpreted \dots}''
\end{quote}

This regime corresponds to a smooth transition in cognitive framing,
where the model increasingly emphasizes relationships and reasoning steps.

\paragraph{High $\alpha$.}
At higher values of $\alpha$, the effect becomes more pronounced but less stable:

\begin{quote}
\textbf{Steered ($\alpha=5$):}  
``\textit{It is important important to interpret interpret the data in a broader
context context of understanding understanding \dots}''
\end{quote}

While the probe-predicted level continues to increase, the output may exhibit
repetition, redundancy, or loss of coherence. This suggests that strong
interventions amplify the level direction beyond the range supported by
the model's training distribution.

\section{Discussion}

Our results show that Bloom-level information is linearly decodable from early layers of multiple LLMs, suggesting that prompts are rapidly organized according to features 
correlated with Bloom levels. The emergence of a consistent Cognitive Separability Onset across architectures further indicates that this organization is not model-specific but reflects a robust representational pattern. Together with the centroid analysis, these findings support the view that the complexity is encoded as progressive geometric separation of adjacent categories within the residual stream.

Importantly, the gap between residual-stream probe accuracy and shallow text baselines provides evidence against the hypothesis 
that probe performance is driven solely by lexical cues such as 
task-indicative verbs. While some lexical signal is present as the non-trivial TF-IDF baseline confirms, the structured, 
ordinal error patterns observed exclusively in residual-stream 
probes suggest that LLMs encode additional information beyond 
what is recoverable from surface form alone.

At the same time, linear separability does not imply causal use: the model may cluster inputs by surface strategy or task type rather than by fully abstract cognitive levels. Although shallow baselines underperform, residual-stream separability could still partially reflect training-induced regularities in prompt formulation. Nonetheless, the structured ordinal error patterns suggest that internal representations preserve meaningful hierarchical relationships rather than arbitrary label boundaries. 
We address this gap directly in Section~\ref{sec:causal}, providing preliminary intervention evidence that the identified representational direction is functionally linked to downstream generation behavior.

\section{Conclusion and Future Work}

Our work provides evidence that LLMs possess a measurable, linearly separable internal representation of complexity that aligns closely with Bloom's Taxonomy. By systematically extracting and analyzing high-dimensional activation vectors from the residual streams of various state-of-the-art models, we have shown evidence that probing performance is not fully explained by surface-level lexical baselines.

Our investigation into the layer-wise dynamics of these representations revealed a striking \textit{Cognitive Separability Onset} occurring in the early-to-intermediate stages of the forward pass. This finding suggests that representations correlated with the 
Bloom information of a task emerge long before the final token 
generation, with representations becoming differentiated in early layers according to Bloom level. Furthermore, the high classification accuracy achieved by our linear probe indicates that these internal representations encode Bloom-level information in a coherent and geometrically structured manner that preserves the ordinal structure of human educational theory.

The validity of these findings is further supported by our control experiments. Shallow text baselines achieve substantially lower accuracy than the residual-stream probes, and their confusion matrices show dispersed, non-ordinal error patterns.
This contrast suggests that while surface phrasing provides a non-trivial baseline signal, the internal representations captured by residual-stream probes exhibit additional structured, ordinal separation across layers.

These results suggest that Bloom's Taxonomy may serve not only as 
an instrument for human learners, but also as a useful lens for 
analyzing internal representations of complexity in LLMs.

Several promising directions emerge from our findings. The causal intervention study presented in Section~\ref{sec:causal} provides initial evidence that Bloom-level representations are functionally active; future work could extend this by selectively ablating individual dimensions most predictive of Bloom level, testing whether targeted feature suppression produces graded degradation consistent with the taxonomy's ordinal structure.

Moreover, cross-lingual and cross-domain analysis would establish the universality of the CSO by extending our methodology to non-English languages and specialized professional domains (e.g., medical diagnosis, legal reasoning).

As the field moves toward increasingly autonomous and complex AI agents, the ability to monitor, interpret, and potentially regulate these settings via their neural correlates will be essential. Our work contributes toward the possibility of evaluating AI against the same sophisticated standards of understanding that we apply to human intelligence.

\section*{Limitations} \label{sec:limitations}
Our analysis focuses exclusively on decoder-only transformer architectures and English educational questions. The generalizability to other architectures (e.g., encoder-only or encoder-decoder models), languages, and non-finetuned base models remains to be established. 
Furthermore, our methodology examines only the residual stream, leaving attention patterns and MLP internals as areas for future investigation. Finally, the causal steering experiments are conducted on a single model; multi-model validation and systematic human evaluation remain open directions.

\appendix
\section{Accuracy per Bloom's Level}\label{app:accuracy_per_level}
\begin{figure*}
    \centering
    \begin{minipage}[b]{0.49\linewidth}
        \centering
        \includegraphics[width=\linewidth]{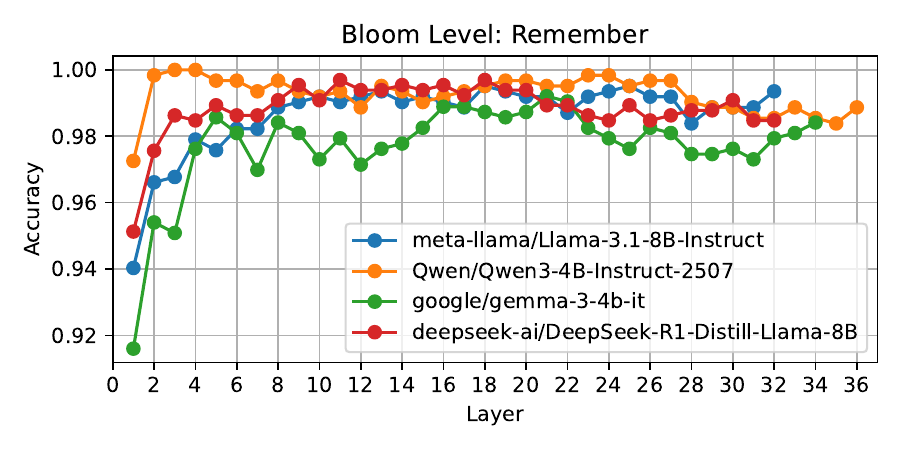}
    \end{minipage}
    \hfill
    \begin{minipage}[b]{0.49\linewidth}
        \centering
        \includegraphics[width=\linewidth]{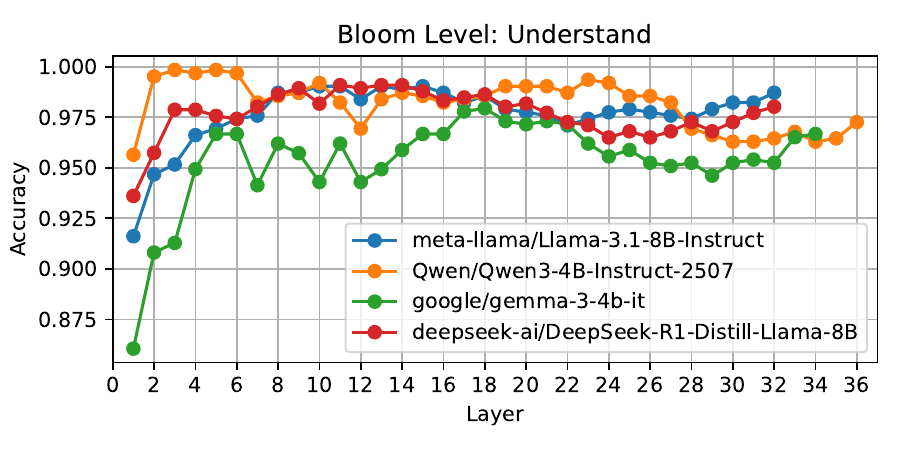}
    \end{minipage}
    \vspace{0.5pt}
    \begin{minipage}[b]{0.49\linewidth}
        \centering
        \includegraphics[width=\linewidth]{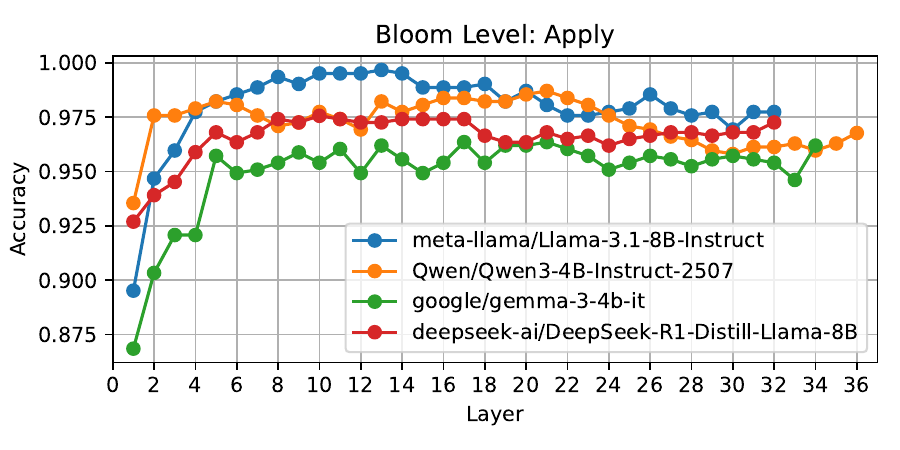}
    \end{minipage}
    \hfill
    \begin{minipage}[b]{0.49\linewidth}
        \centering
        \includegraphics[width=\linewidth]{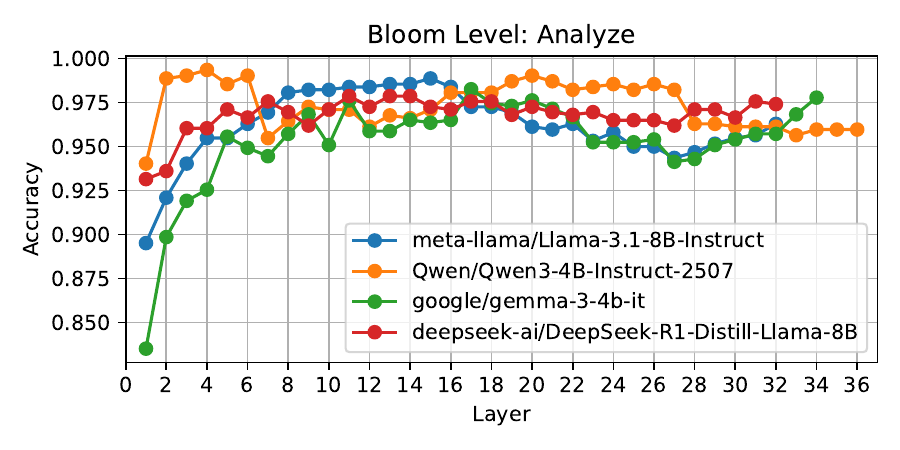}
    \end{minipage}
    \vspace{0.5pt}
    \begin{minipage}[b]{0.49\linewidth}
        \centering
        \includegraphics[width=\linewidth]{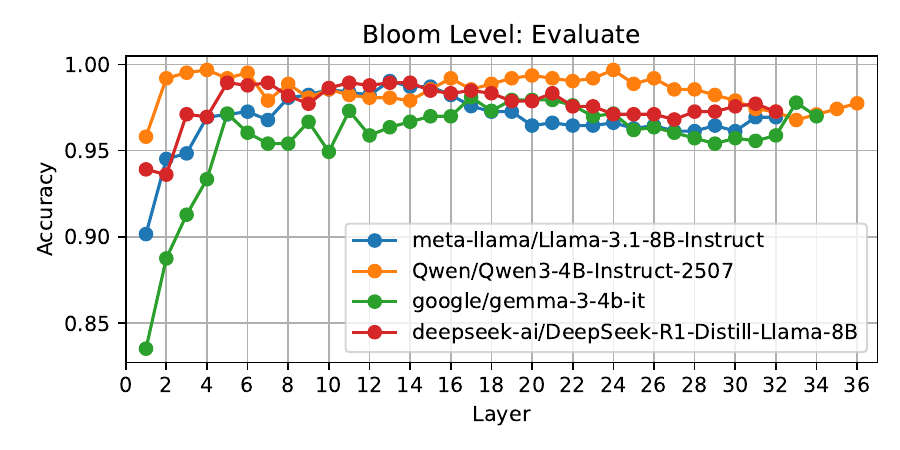}
    \end{minipage}
    \hfill
    \begin{minipage}[b]{0.49\linewidth}
        \centering
        \includegraphics[width=\linewidth]{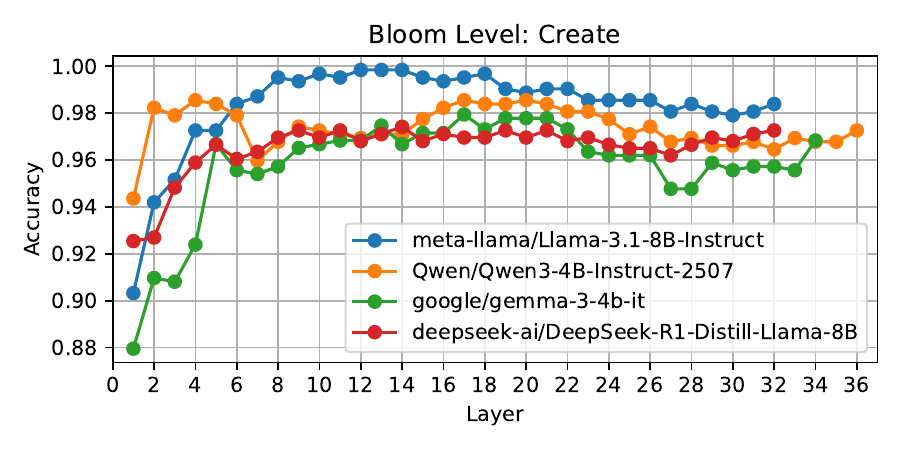}
    \end{minipage}

    \caption{Layer-wise linear probe accuracy for each Bloom level.}
    \label{fig:accuracy_bloom_levels}
\end{figure*}
Figure~\ref{fig:accuracy_bloom_levels} reports the probe accuracy separately for each Bloom level across layers and models. 
A consistent pattern emerges across architectures: all levels converge to near-ceiling performance by the CSO layer 
$l^\star=5$.

\section{Confusion Matrix}\label{app:confusion_matrix}
\begin{figure*}
    \centering
    \begin{minipage}[b]{0.3\textwidth}
        \centering
        \includegraphics[width=\linewidth]{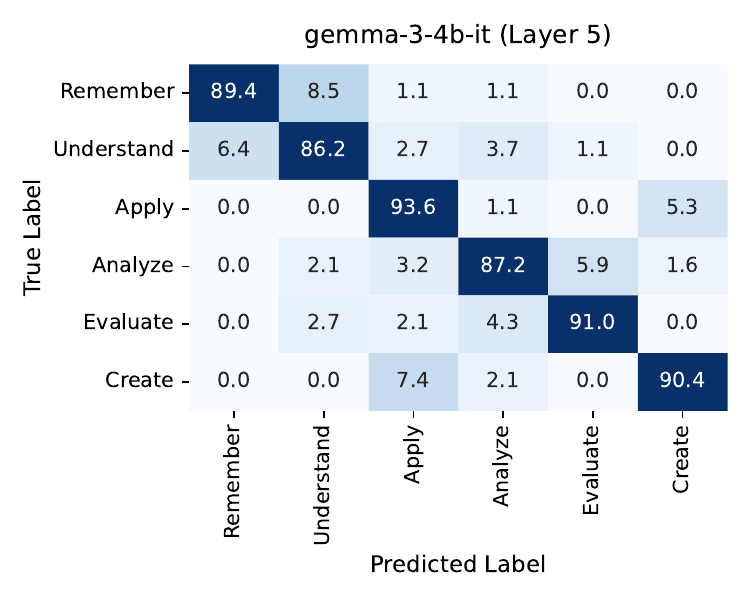}
    \end{minipage}
    \hfill
    \begin{minipage}[b]{0.3\textwidth}
        \centering
        \includegraphics[width=\linewidth]{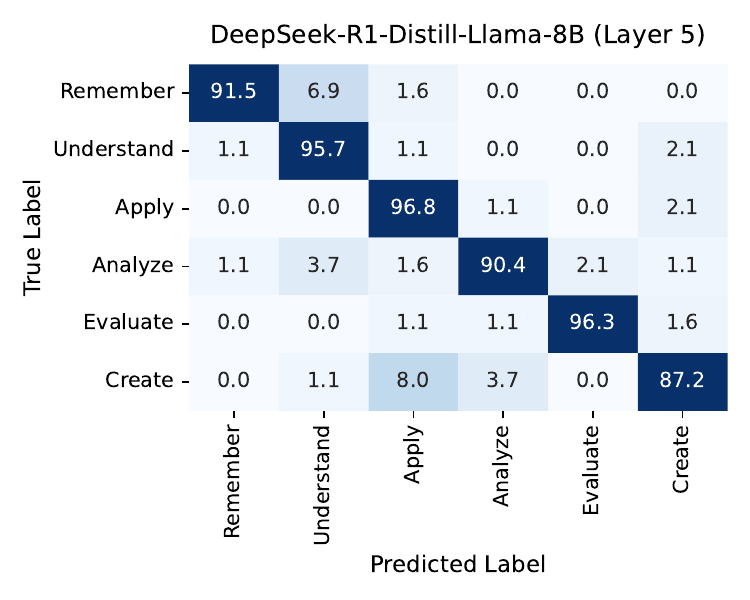}
    \end{minipage}
    \hfill
    \begin{minipage}[b]{0.3\textwidth}
        \centering
        \includegraphics[width=\linewidth]{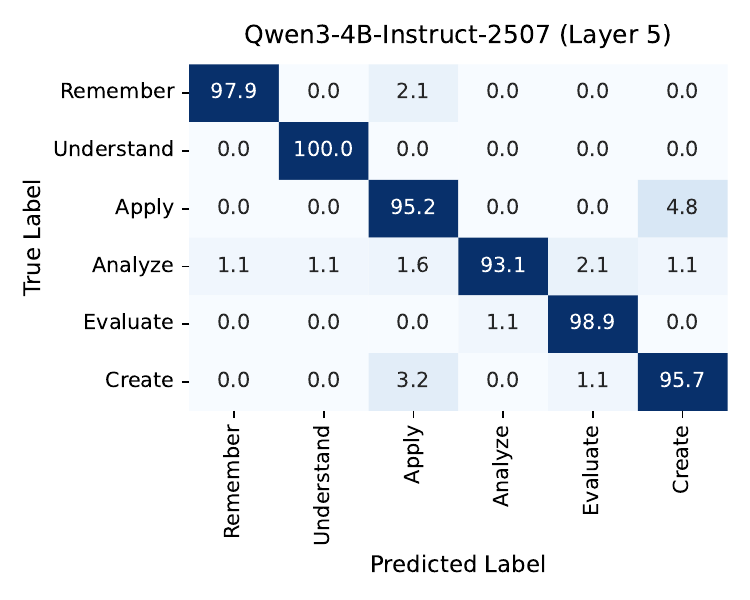}
    \end{minipage}

    \caption{Confusion matrices of the linear probe for three representative models at the CSO layer $l^\star=5$: \texttt{gemma-3-4b-it}, \texttt{DeepSeek-R1-Distill-Llama-8B}, and \texttt{Qwen3-4B-Instruct-2507}.}
    \label{fig:confusion_matrices_all}
\end{figure*}
The confusion matrices in Figure~\ref{fig:confusion_matrices_all} confirm that classification errors are highly structured 
rather than random. Across all architectures, misclassifications predominantly occur between 
adjacent Bloom levels (e.g., Analyze vs.\ Evaluate). Extreme long-range confusions (such as Remember vs. Create) are consistently absent, though a localized non-adjacent confusion occurs between Apply and Create probably due to shared imperative prompt phrasing.

This adjacency bias provides further evidence that complexity is encoded within the representation space.
The observed error concentration along the sub-diagonals indicates that misclassifications predominantly occur between neighboring Bloom categories, suggesting a locally structured representational boundary rather than arbitrary classification.
The consistency of 
this pattern across models suggests that it reflects a general property of Transformer-based 
representations rather than an artifact of a specific architecture.

\section{Centroids}
\label{app:centroids}
\begin{figure*}
    \centering
    \begin{minipage}[b]{0.3\textwidth}
        \centering
        \includegraphics[width=\linewidth]{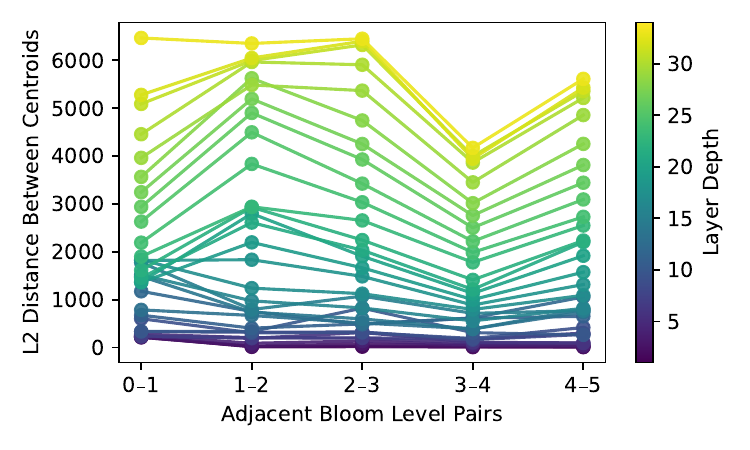}
    \end{minipage}
    \hfill
    \begin{minipage}[b]{0.3\textwidth}
        \centering
        \includegraphics[width=\linewidth]{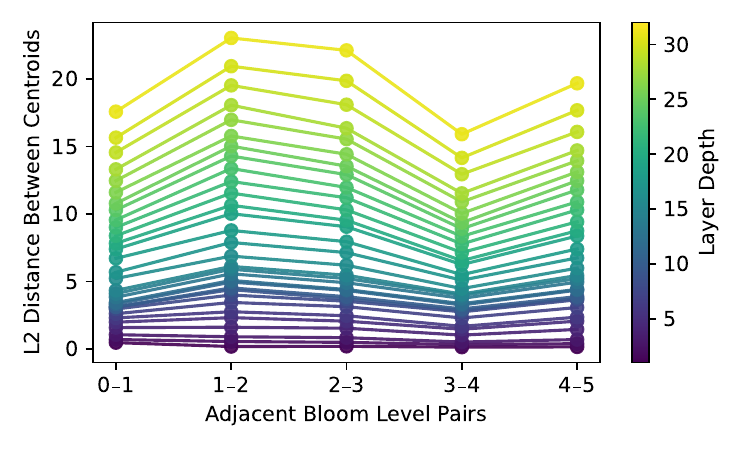}
    \end{minipage}
    \hfill
    \begin{minipage}[b]{0.3\textwidth}
        \centering
        \includegraphics[width=\linewidth]{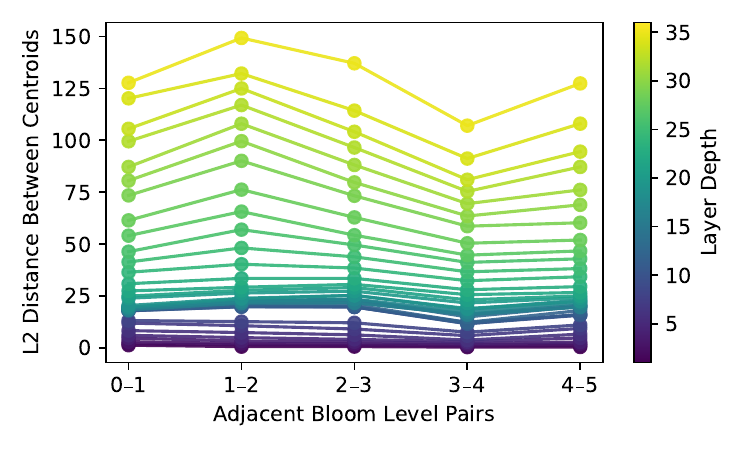}
    \end{minipage}

    \caption{Layer-wise Euclidean distances between adjacent Bloom-level centroids for all evaluated models: \texttt{gemma-3-4b-it}, \texttt{DeepSeek-R1-Distill-Llama-8B}, and \texttt{Qwen3-4B-Instruct-2507}.}
    \label{fig:centroids_all}
\end{figure*}
Figure~\ref{fig:centroids_all} visualizes the Euclidean distance between adjacent Bloom-level centroids across layers 
for all evaluated models. In early layers, centroid distances are small, indicating that representations of different cognitive levels are highly entangled. 

Starting around the CSO layer, distances increase sharply and then continue to expand 
monotonically with depth. This progressive separation demonstrates that deeper layers do 
not merely preserve linear separability, but actively amplify geometric distinctions between 
cognitive categories.

Importantly, this analysis is probe-independent and therefore confirms that the Cognitive Separability Onset corresponds to an intrinsic restructuring of the latent space rather than an artifact of linear classification.

\section*{Acknowledgements}
This research did not receive any specific grant from funding agencies in the public, commercial, or not-for-profit sectors.
The authors declare no competing interests.

\printcredits

\bibliographystyle{cas-model2-names}

\bibliography{cas-refs}

\begin{thebibliography}{36}
\expandafter\ifx\csname natexlab\endcsname\relax\def\natexlab#1{#1}\fi
\providecommand{\url}[1]{\texttt{#1}}
\providecommand{\href}[2]{#2}
\providecommand{\path}[1]{#1}
\providecommand{\DOIprefix}{doi:}
\providecommand{\ArXivprefix}{arXiv:}
\providecommand{\URLprefix}{URL: }
\providecommand{\Pubmedprefix}{pmid:}
\providecommand{\doi}[1]{\href{http://dx.doi.org/#1}{\path{#1}}}
\providecommand{\Pubmed}[1]{\href{pmid:#1}{\path{#1}}}
\providecommand{\bibinfo}[2]{#2}
\ifx\xfnm\relax \def\xfnm[#1]{\unskip,\space#1}\fi
\bibitem[{Anderson and Krathwohl(2001)}]{anderson2001taxonomy}
\bibinfo{author}{Anderson, L.W.}, \bibinfo{author}{Krathwohl, D.R.}, \bibinfo{year}{2001}.
\newblock \bibinfo{title}{A taxonomy for learning, teaching, and assessing: A revision of Bloom's taxonomy of educational objectives: complete edition}.
\newblock \bibinfo{publisher}{Addison Wesley Longman, Inc.}
\bibitem[{Belinkov(2022)}]{belinkov2022probing}
\bibinfo{author}{Belinkov, Y.}, \bibinfo{year}{2022}.
\newblock \bibinfo{title}{Probing classifiers: Promises, shortcomings, and advances}.
\newblock \bibinfo{journal}{Computational Linguistics} \bibinfo{volume}{48}, \bibinfo{pages}{207--219}.
\bibitem[{Benito-Rodriguez et~al.(2025)Benito-Rodriguez, Urdshals, Nasufi and Pochinkov}]{benito2025beyond}
\bibinfo{author}{Benito-Rodriguez, {\'E}.}, \bibinfo{author}{Urdshals, E.}, \bibinfo{author}{Nasufi, J.}, \bibinfo{author}{Pochinkov, N.}, \bibinfo{year}{2025}.
\newblock \bibinfo{title}{Beyond tokens in language models: Interpreting activations through text genre chunks}.
\newblock \bibinfo{journal}{arXiv preprint arXiv:2511.16540} .
\bibitem[{Bereska and Gavves(2024)}]{bereska2024mechanistic}
\bibinfo{author}{Bereska, L.F.}, \bibinfo{author}{Gavves, E.}, \bibinfo{year}{2024}.
\newblock \bibinfo{title}{Mechanistic interpretability for ai safety — a review}.
\newblock \bibinfo{journal}{TMLR} \href{http://arxiv.org/abs/2404.14082}{\tt arXiv:2404.14082}.
\bibitem[{Brown et~al.(2020)Brown, Mann, Ryder, Subbiah, Kaplan, Dhariwal, Neelakantan, Shyam, Sastry, Askell et~al.}]{brown2020language}
\bibinfo{author}{Brown, T.}, \bibinfo{author}{Mann, B.}, \bibinfo{author}{Ryder, N.}, \bibinfo{author}{Subbiah, M.}, \bibinfo{author}{Kaplan, J.D.}, \bibinfo{author}{Dhariwal, P.}, \bibinfo{author}{Neelakantan, A.}, \bibinfo{author}{Shyam, P.}, \bibinfo{author}{Sastry, G.}, \bibinfo{author}{Askell, A.}, et~al., \bibinfo{year}{2020}.
\newblock \bibinfo{title}{Language models are few-shot learners}.
\newblock \bibinfo{journal}{Advances in neural information processing systems} \bibinfo{volume}{33}, \bibinfo{pages}{1877--1901}.
\bibitem[{Budagam et~al.(2024)Budagam, Kumar, Khoshnoodi, KJ, Jain and Chadha}]{budagam2024hierarchical}
\bibinfo{author}{Budagam, D.}, \bibinfo{author}{Kumar, A.}, \bibinfo{author}{Khoshnoodi, M.}, \bibinfo{author}{KJ, S.}, \bibinfo{author}{Jain, V.}, \bibinfo{author}{Chadha, A.}, \bibinfo{year}{2024}.
\newblock \bibinfo{title}{Hierarchical prompting taxonomy: A universal evaluation framework for large language models aligned with human cognitive principles}.
\newblock \bibinfo{journal}{arXiv preprint arXiv:2406.12644} .
\bibitem[{Cywi{\'n}ski et~al.(2025)Cywi{\'n}ski, Ryd, Rajamanoharan and Nanda}]{cywinski2025towards}
\bibinfo{author}{Cywi{\'n}ski, B.}, \bibinfo{author}{Ryd, E.}, \bibinfo{author}{Rajamanoharan, S.}, \bibinfo{author}{Nanda, N.}, \bibinfo{year}{2025}.
\newblock \bibinfo{title}{Towards eliciting latent knowledge from llms with mechanistic interpretability}.
\newblock \bibinfo{journal}{arXiv preprint arXiv:2505.14352} .
\bibitem[{Elhage et~al.(2021)Elhage, Nanda, Olsson, Henighan, Joseph, Mann, Askell, Bai, Chen, Conerly et~al.}]{elhage2021mathematical}
\bibinfo{author}{Elhage, N.}, \bibinfo{author}{Nanda, N.}, \bibinfo{author}{Olsson, C.}, \bibinfo{author}{Henighan, T.}, \bibinfo{author}{Joseph, N.}, \bibinfo{author}{Mann, B.}, \bibinfo{author}{Askell, A.}, \bibinfo{author}{Bai, Y.}, \bibinfo{author}{Chen, A.}, \bibinfo{author}{Conerly, T.}, et~al., \bibinfo{year}{2021}.
\newblock \bibinfo{title}{A mathematical framework for transformer circuits}.
\newblock \bibinfo{journal}{Transformer Circuits Thread} \bibinfo{volume}{1}, \bibinfo{pages}{12}.
\bibitem[{Elkins et~al.(2024)Elkins, Kochmar, Cheung and Serban}]{elkins2024teachers}
\bibinfo{author}{Elkins, S.}, \bibinfo{author}{Kochmar, E.}, \bibinfo{author}{Cheung, J.C.}, \bibinfo{author}{Serban, I.}, \bibinfo{year}{2024}.
\newblock \bibinfo{title}{How teachers can use large language models and bloom’s taxonomy to create educational quizzes}, in: \bibinfo{booktitle}{Proceedings of the AAAI Conference on Artificial Intelligence}, pp. \bibinfo{pages}{23084--23091}.
\bibitem[{Gantla(2025)}]{gantla2025exploring}
\bibinfo{author}{Gantla, S.R.}, \bibinfo{year}{2025}.
\newblock \bibinfo{title}{Exploring mechanistic interpretability in large language models: Challenges, approaches, and insights}, in: \bibinfo{booktitle}{2025 International Conference on Data Science, Agents \& Artificial Intelligence (ICDSAAI)}, \bibinfo{organization}{IEEE}. pp. \bibinfo{pages}{1--8}.
\bibitem[{Hadifar et~al.(2023)Hadifar, Bitew, Deleu, Develder and Demeester}]{hadifar2023eduqg}
\bibinfo{author}{Hadifar, A.}, \bibinfo{author}{Bitew, S.K.}, \bibinfo{author}{Deleu, J.}, \bibinfo{author}{Develder, C.}, \bibinfo{author}{Demeester, T.}, \bibinfo{year}{2023}.
\newblock \bibinfo{title}{Eduqg: A multi-format multiple-choice dataset for the educational domain}.
\newblock \bibinfo{journal}{Ieee Access} \bibinfo{volume}{11}, \bibinfo{pages}{20885--20896}.
\bibitem[{Hendrycks et~al.(2021)Hendrycks, Burns, Basart, Zou, Mazeika, Song and Steinhardt}]{hendrycks2020measuring}
\bibinfo{author}{Hendrycks, D.}, \bibinfo{author}{Burns, C.}, \bibinfo{author}{Basart, S.}, \bibinfo{author}{Zou, A.}, \bibinfo{author}{Mazeika, M.}, \bibinfo{author}{Song, D.}, \bibinfo{author}{Steinhardt, J.}, \bibinfo{year}{2021}.
\newblock \bibinfo{title}{Measuring massive multitask language understanding}, in: \bibinfo{booktitle}{International Conference on Learning Representations}.
\newblock \URLprefix \url{https://openreview.net/forum?id=d7KBjmI3GmQ}.
\bibitem[{Herrmann-Werner et~al.(2024)Herrmann-Werner, Festl-Wietek, Holderried, Herschbach, Griewatz, Masters, Zipfel and Mahling}]{herrmann2024assessing}
\bibinfo{author}{Herrmann-Werner, A.}, \bibinfo{author}{Festl-Wietek, T.}, \bibinfo{author}{Holderried, F.}, \bibinfo{author}{Herschbach, L.}, \bibinfo{author}{Griewatz, J.}, \bibinfo{author}{Masters, K.}, \bibinfo{author}{Zipfel, S.}, \bibinfo{author}{Mahling, M.}, \bibinfo{year}{2024}.
\newblock \bibinfo{title}{Assessing chatgpt’s mastery of bloom’s taxonomy using psychosomatic medicine exam questions: mixed-methods study}.
\newblock \bibinfo{journal}{Journal of medical Internet research} \bibinfo{volume}{26}, \bibinfo{pages}{e52113}.
\bibitem[{Hewitt and Liang(2019)}]{hewitt2019designing}
\bibinfo{author}{Hewitt, J.}, \bibinfo{author}{Liang, P.}, \bibinfo{year}{2019}.
\newblock \bibinfo{title}{Designing and interpreting probes with control tasks}, in: \bibinfo{editor}{Inui, K.}, \bibinfo{editor}{Jiang, J.}, \bibinfo{editor}{Ng, V.}, \bibinfo{editor}{Wan, X.} (Eds.), \bibinfo{booktitle}{Proceedings of the 2019 Conference on Empirical Methods in Natural Language Processing and the 9th International Joint Conference on Natural Language Processing (EMNLP-IJCNLP)}, \bibinfo{publisher}{Association for Computational Linguistics}, \bibinfo{address}{Hong Kong, China}. pp. \bibinfo{pages}{2733--2743}.
\newblock \URLprefix \url{https://aclanthology.org/D19-1275/}, \DOIprefix\doi{10.18653/v1/D19-1275}.
\bibitem[{Hmoud and Ali(2024)}]{hmoud2024aied}
\bibinfo{author}{Hmoud, M.}, \bibinfo{author}{Ali, S.}, \bibinfo{year}{2024}.
\newblock \bibinfo{title}{Aied bloom’s taxonomy: A proposed model for enhancing educational efficiency and effectiveness in the artificial intelligence era}.
\newblock \bibinfo{journal}{The International Journal of Technologies in Learning} \bibinfo{volume}{31}, \bibinfo{pages}{111}.
\bibitem[{Huber and Niklaus(2025)}]{huber2025llms}
\bibinfo{author}{Huber, T.}, \bibinfo{author}{Niklaus, C.}, \bibinfo{year}{2025}.
\newblock \bibinfo{title}{Llms meet bloom’s taxonomy: A cognitive view on large language model evaluations}, in: \bibinfo{booktitle}{Proceedings of the 31st International Conference on Computational Linguistics}, pp. \bibinfo{pages}{5211--5246}.
\bibitem[{Jankowski et~al.(2026)Jankowski, Radicchi, Serrano, Bogu{\~n}{\'a} and Fortunato}]{jankowski2025task}
\bibinfo{author}{Jankowski, R.}, \bibinfo{author}{Radicchi, F.}, \bibinfo{author}{Serrano, M.{\'A}.}, \bibinfo{author}{Bogu{\~n}{\'a}, M.}, \bibinfo{author}{Fortunato, S.}, \bibinfo{year}{2026}.
\newblock \bibinfo{title}{Task complexity shapes internal representations and robustness in neural networks}.
\newblock \bibinfo{journal}{Machine Learning: Science and Technology} \bibinfo{volume}{7}, \bibinfo{pages}{025045}.
\bibitem[{Jin et~al.(2025)Jin, Yu, Huang, Zeng, Wang, Hua, Zhao, Mei, Meng, Ding et~al.}]{jin2025exploring}
\bibinfo{author}{Jin, M.}, \bibinfo{author}{Yu, Q.}, \bibinfo{author}{Huang, J.}, \bibinfo{author}{Zeng, Q.}, \bibinfo{author}{Wang, Z.}, \bibinfo{author}{Hua, W.}, \bibinfo{author}{Zhao, H.}, \bibinfo{author}{Mei, K.}, \bibinfo{author}{Meng, Y.}, \bibinfo{author}{Ding, K.}, et~al., \bibinfo{year}{2025}.
\newblock \bibinfo{title}{Exploring concept depth: How large language models acquire knowledge and concept at different layers?}, in: \bibinfo{booktitle}{Proceedings of the 31st international conference on computational linguistics}, pp. \bibinfo{pages}{558--573}.
\bibitem[{Kim et~al.(2018)Kim, Wattenberg, Gilmer, Cai, Wexler, Viegas et~al.}]{kim2018interpretability}
\bibinfo{author}{Kim, B.}, \bibinfo{author}{Wattenberg, M.}, \bibinfo{author}{Gilmer, J.}, \bibinfo{author}{Cai, C.}, \bibinfo{author}{Wexler, J.}, \bibinfo{author}{Viegas, F.}, et~al., \bibinfo{year}{2018}.
\newblock \bibinfo{title}{Interpretability beyond feature attribution: Quantitative testing with concept activation vectors (tcav)}, in: \bibinfo{booktitle}{International conference on machine learning}, \bibinfo{organization}{PMLR}. pp. \bibinfo{pages}{2668--2677}.
\bibitem[{Kim et~al.(2025)Kim, Evans and Schein}]{kim2025linear}
\bibinfo{author}{Kim, J.}, \bibinfo{author}{Evans, J.}, \bibinfo{author}{Schein, A.}, \bibinfo{year}{2025}.
\newblock \bibinfo{title}{Linear representations of political perspective emerge in large language models}, in: \bibinfo{booktitle}{The Thirteenth International Conference on Learning Representations}.
\newblock \URLprefix \url{https://openreview.net/forum?id=rwqShzb9li}.
\bibitem[{Krathwohl(2002)}]{krathwohl2002revision}
\bibinfo{author}{Krathwohl, D.R.}, \bibinfo{year}{2002}.
\newblock \bibinfo{title}{A revision of bloom's taxonomy: An overview}.
\newblock \bibinfo{journal}{Theory into practice} \bibinfo{volume}{41}, \bibinfo{pages}{212--218}.
\bibitem[{Kumar et~al.(2025)Kumar, Gulwani and Singh}]{kumar2025automated}
\bibinfo{author}{Kumar, R.}, \bibinfo{author}{Gulwani, D.}, \bibinfo{author}{Singh, S.}, \bibinfo{year}{2025}.
\newblock \bibinfo{title}{Automated analysis of learning outcomes and exam questions based on bloom's taxonomy}.
\newblock \bibinfo{journal}{arXiv preprint arXiv:2511.10903} .
\bibitem[{Luo et~al.(2025a)Luo, Liu, Pang, McKay, Chen, Buchanan and Chang}]{luo2025enhanced}
\bibinfo{author}{Luo, Y.}, \bibinfo{author}{Liu, T.}, \bibinfo{author}{Pang, P.C.I.}, \bibinfo{author}{McKay, D.}, \bibinfo{author}{Chen, Z.}, \bibinfo{author}{Buchanan, G.}, \bibinfo{author}{Chang, S.}, \bibinfo{year}{2025}a.
\newblock \bibinfo{title}{Enhanced bloom's educational taxonomy for fostering information literacy in the era of large language models}.
\newblock \bibinfo{journal}{arXiv preprint arXiv:2503.19434} .
\bibitem[{Luo et~al.(2025b)Luo, Zhou and Dong}]{luo2025inversescope}
\bibinfo{author}{Luo, Y.}, \bibinfo{author}{Zhou, Z.}, \bibinfo{author}{Dong, B.}, \bibinfo{year}{2025}b.
\newblock \bibinfo{title}{Inversescope: Scalable activation inversion for interpreting large language models}.
\newblock \bibinfo{journal}{arXiv preprint arXiv:2506.07406} .
\bibitem[{Olah et~al.(2020)Olah, Cammarata, Schubert, Goh, Petrov and Carter}]{olah2020zoom}
\bibinfo{author}{Olah, C.}, \bibinfo{author}{Cammarata, N.}, \bibinfo{author}{Schubert, L.}, \bibinfo{author}{Goh, G.}, \bibinfo{author}{Petrov, M.}, \bibinfo{author}{Carter, S.}, \bibinfo{year}{2020}.
\newblock \bibinfo{title}{Zoom in: An introduction to circuits}.
\newblock \bibinfo{journal}{Distill} \bibinfo{volume}{5}, \bibinfo{pages}{e00024--001}.
\bibitem[{Qaiser and Ali(2018)}]{qaiser2018text}
\bibinfo{author}{Qaiser, S.}, \bibinfo{author}{Ali, R.}, \bibinfo{year}{2018}.
\newblock \bibinfo{title}{Text mining: use of tf-idf to examine the relevance of words to documents}.
\newblock \bibinfo{journal}{International journal of computer applications} \bibinfo{volume}{181}, \bibinfo{pages}{25--29}.
\bibitem[{Rai et~al.(2024)Rai, Zhou, Feng, Saparov and Yao}]{rai2024practical}
\bibinfo{author}{Rai, D.}, \bibinfo{author}{Zhou, Y.}, \bibinfo{author}{Feng, S.}, \bibinfo{author}{Saparov, A.}, \bibinfo{author}{Yao, Z.}, \bibinfo{year}{2024}.
\newblock \bibinfo{title}{A practical review of mechanistic interpretability for transformer-based language models}.
\newblock \bibinfo{journal}{arXiv preprint arXiv:2407.02646} .
\bibitem[{Raimondi et~al.(2025)Raimondi, Dalbagno and Gabbrielli}]{raimondi2025analysing}
\bibinfo{author}{Raimondi, B.}, \bibinfo{author}{Dalbagno, D.}, \bibinfo{author}{Gabbrielli, M.}, \bibinfo{year}{2025}.
\newblock \bibinfo{title}{Analysing moral bias in finetuned llms through mechanistic interpretability}.
\newblock \bibinfo{journal}{arXiv preprint arXiv:2510.12229} .
\bibitem[{Reimers and Gurevych(2019)}]{reimers-2019-sentence-bert}
\bibinfo{author}{Reimers, N.}, \bibinfo{author}{Gurevych, I.}, \bibinfo{year}{2019}.
\newblock \bibinfo{title}{Sentence-bert: Sentence embeddings using siamese bert-networks}, in: \bibinfo{booktitle}{Proceedings of the 2019 Conference on Empirical Methods in Natural Language Processing}, \bibinfo{publisher}{Association for Computational Linguistics}.
\newblock \URLprefix \url{https://arxiv.org/abs/1908.10084}.
\bibitem[{Shu et~al.(2025)Shu, Wu, Zhao, Rai, Yao, Liu and Du}]{shu2025survey}
\bibinfo{author}{Shu, D.}, \bibinfo{author}{Wu, X.}, \bibinfo{author}{Zhao, H.}, \bibinfo{author}{Rai, D.}, \bibinfo{author}{Yao, Z.}, \bibinfo{author}{Liu, N.}, \bibinfo{author}{Du, M.}, \bibinfo{year}{2025}.
\newblock \bibinfo{title}{A survey on sparse autoencoders: Interpreting the internal mechanisms of large language models}, in: \bibinfo{editor}{Christodoulopoulos, C.}, \bibinfo{editor}{Chakraborty, T.}, \bibinfo{editor}{Rose, C.}, \bibinfo{editor}{Peng, V.} (Eds.), \bibinfo{booktitle}{Findings of the Association for Computational Linguistics: EMNLP 2025}, \bibinfo{publisher}{Association for Computational Linguistics}, \bibinfo{address}{Suzhou, China}. pp. \bibinfo{pages}{1690--1712}.
\newblock \URLprefix \url{https://aclanthology.org/2025.findings-emnlp.89/}, \DOIprefix\doi{10.18653/v1/2025.findings-emnlp.89}.
\bibitem[{Simbeck and Mahran(2025)}]{simbeck2025mechanistic}
\bibinfo{author}{Simbeck, K.}, \bibinfo{author}{Mahran, M.}, \bibinfo{year}{2025}.
\newblock \bibinfo{title}{Mechanistic interpretability with saes: Probing religion, violence, and geography in large language models}.
\newblock \bibinfo{journal}{Proceedings of the 3rd Workshop on Fairness and Bias in AI} .
\bibitem[{Vaswani et~al.(2017)Vaswani, Shazeer, Parmar, Uszkoreit, Jones, Gomez, Kaiser and Polosukhin}]{vaswani2017attention}
\bibinfo{author}{Vaswani, A.}, \bibinfo{author}{Shazeer, N.}, \bibinfo{author}{Parmar, N.}, \bibinfo{author}{Uszkoreit, J.}, \bibinfo{author}{Jones, L.}, \bibinfo{author}{Gomez, A.N.}, \bibinfo{author}{Kaiser, {\L}.}, \bibinfo{author}{Polosukhin, I.}, \bibinfo{year}{2017}.
\newblock \bibinfo{title}{Attention is all you need}.
\newblock \bibinfo{journal}{Advances in neural information processing systems} \bibinfo{volume}{30}.
\bibitem[{Yu et~al.(2026)Yu, Wu, Lin and Lobczowski}]{yu2025think}
\bibinfo{author}{Yu, Y.}, \bibinfo{author}{Wu, M.}, \bibinfo{author}{Lin, Y.}, \bibinfo{author}{Lobczowski, N.G.}, \bibinfo{year}{2026}.
\newblock \bibinfo{title}{Think: Can large language models think-aloud?}, in: \bibinfo{booktitle}{International Conference on Artificial Intelligence in Education}, \bibinfo{organization}{Springer}. pp. \bibinfo{pages}{509--517}.
\bibitem[{Zaman et~al.(2024)Zaman, Islam, Islam and Sayed}]{zaman2024dataset}
\bibinfo{author}{Zaman, K.A.U.}, \bibinfo{author}{Islam, A.}, \bibinfo{author}{Islam, Y.M.}, \bibinfo{author}{Sayed, M.A.}, \bibinfo{year}{2024}.
\newblock \bibinfo{title}{Dataset of computer science course queries from students: Categorized and scored according to bloom's taxonomy}.
\newblock \bibinfo{journal}{Data in Brief} \bibinfo{volume}{53}, \bibinfo{pages}{110109}.
\bibitem[{Zhao et~al.(2024)Zhao, Yang, Shen, Lakkaraju and Du}]{zhao2024towards}
\bibinfo{author}{Zhao, H.}, \bibinfo{author}{Yang, F.}, \bibinfo{author}{Shen, B.}, \bibinfo{author}{Lakkaraju, H.}, \bibinfo{author}{Du, M.}, \bibinfo{year}{2024}.
\newblock \bibinfo{title}{Towards uncovering how large language model works: An explainability perspective}.
\newblock \bibinfo{journal}{arXiv preprint arXiv:2402.10688} .
\bibitem[{Zoumpoulidi et~al.(2025)Zoumpoulidi, Paraskevopoulos and Potamianos}]{zoumpoulidi2025bloomwise}
\bibinfo{author}{Zoumpoulidi, M.E.}, \bibinfo{author}{Paraskevopoulos, G.}, \bibinfo{author}{Potamianos, A.}, \bibinfo{year}{2025}.
\newblock \bibinfo{title}{Bloomwise: enhancing problem-solving capabilities of large language models using bloom’s-taxonomy-inspired prompts}, in: \bibinfo{booktitle}{Proceedings of The 3rd Workshop on Mathematical Natural Language Processing (MathNLP 2025)}, pp. \bibinfo{pages}{34--49}.

\end{thebibliography}

\vskip3pt
\newpage
\bio{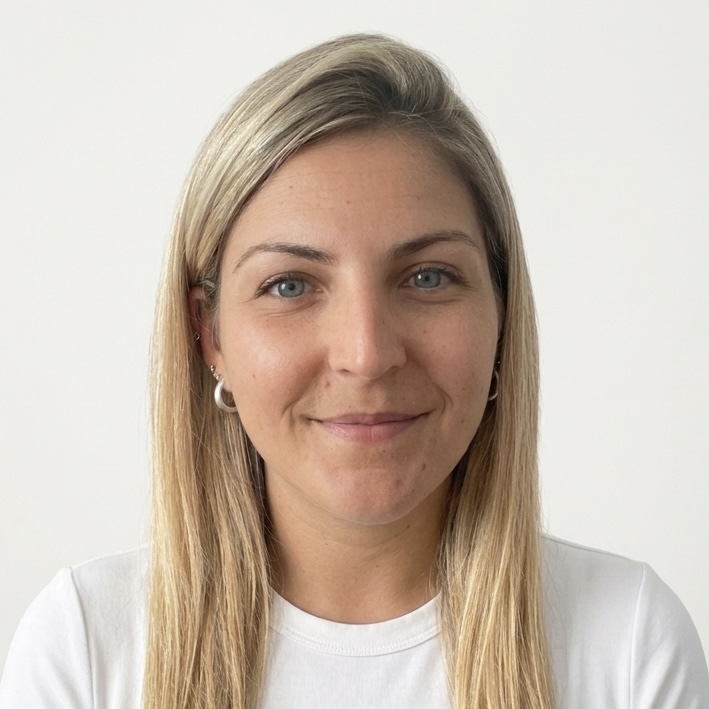}
Bianca Raimondi holds a Master's degree in Computer Science from the University of Bologna. She is currently a PhD student specialising in Data Science and Computation. Her research focuses on applying Large Language Models in education, particularly examining the biases of these models and how they represent information internally.
\endbio

\bio{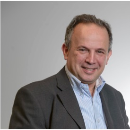}
Maurizio Gabbrielli is professor of Computer Science since 2001 at the Department of Computer Science and Engineering of the University of Bologna and Associate dean for AI at Bologna Business School. He has been Head of the Department of Computer Science and Engineering  and member of the INRIA project team FOCUS. He received his Ph.d. in Computer Science in 1992 from the University of Pisa and has been employed at Centrum Wiskunde \& Informatica (CWI, Amsterdam), at the University of Pisa and at the University of Udine.
\endbio

\end{document}